# Towards Real-Time 2D Mapping: Harnessing Drones, AI, and Computer Vision for Advanced Insights

Bharath Kumar Agnur

Department of Electronics and Communication Engineering

Vidya Jyothi Institute of Technology

Aziz Nagar, Hyderabad, Telangana, India - 500075

agnur.bharath@gmail.com

**Abstract**

Real-time 2D mapping is a vital tool in aerospace and defense, where accurate and timely geographic data is essential for operations like surveillance, reconnaissance, and target tracking. This project introduces a cutting-edge mapping system that integrates drone imagery with machine learning and computer vision to address challenges in processing speed, accuracy, and adaptability to diverse terrains. By automating feature detection, image matching, and stitching, the system generates seamless, high-resolution maps with minimal delay, providing strategic advantages in defense operations.

Implemented in Python, the system leverages OpenCV for image processing, NumPy for efficient computations, and Concurrent.futures for parallel processing. ORB (Oriented FAST and Rotated BRIEF) handles feature detection, while FLANN (Fast Library for Approximate Nearest Neighbors) ensures precise keypoint matching. Homography transformations align overlapping images, creating distortion-free maps in real time. This automated approach eliminates manual intervention, enabling live updates critical in dynamic environments. Designed for adaptability, the system performs well under varying light conditions and rugged terrains, making it highly effective in aerospace and defense scenarios. Testing demonstrates significant improvements in speed and accuracy compared to traditional methods, enhancing situational awareness and decision-making. This scalable solution leverages advanced technologies to deliver reliable, actionable data for mission-critical operations.

**Keywords:** 2D Mapping; Drone Image; Machine Learning; Computer Vision; UAV

## 1. Introduction

In today's world, geographic data plays a critical role in various industries, enabling informed decision-making and efficient management of resources. Real-time mapping has become essential in domains such as environmental monitoring, urban planning, disaster management, and agriculture. It provides stakeholders with up-to-date visual representations of terrains, aiding in applications such as infrastructure development, emergency response, and precision farming. However, the complexity of these applications demands advanced solutions capable of handling the challenges posed by traditional mapping methods. Conventional mapping systems often rely on satellite imagery or manual surveying, which can be time-consuming, expensive, and limited in accuracy. They struggle with real-time data processing, are prone to delays, and lack the adaptability needed to address the dynamic nature of diverse terrains. Additionally, these systems often require significant human intervention for feature identification, image alignment, and error correction. Such limitations make traditional methods unsuitable for scenarios that demand immediate results and high precision, such as disaster relief or rapidly changing environmental conditions. The advent of drone technology has revolutionized the collection of high-resolution aerial imagery, providing an efficient and flexible alternative to traditional data acquisition techniques. Drones can quickly cover large areas, capture detailed images, and access hard to-reach locations. However, the utility of drone imagery depends on how effectively the data is processed and converted into actionable insights. Processing raw images into seamless maps requires addressing several challenges, including feature detection, keypoint matching, image alignment, and stitching accuracy. These processes must also be optimized for real-time performance to meet the demands of critical applications. This project aims to address these challenges by developing a real-time 2D mapping system that leverages advanced machine learning and computer vision techniques. The system automates the mapping workflow, integrating state-of-the-art algorithms for feature detection, image matching, and seamless stitching. By utilizing drone imagery as input, the system efficiently processes high-resolution images to generate continuous, distortion-free maps in real time. It employs ORB (Oriented FAST and Rotated BRIEF) for feature detection, FLANN (Fast Library for Approximate Nearest Neighbors) for keypoint matching, and homography transformations to align overlapping images. The workflow is implemented in Python, taking advantage of powerful libraries such as OpenCV, NumPy, and Concurrent.futures for optimized performance. A key innovation of the system is its adaptability to diverse terrains and conditions. Whether mapping urban landscapes, agricultural fields, or rugged terrains, the system's robust algorithms ensure high accuracy and reliability. Real-time feedback during the stitching process provides immediate updates, making the system user-friendly and efficient. Furthermore, the automated workflow eliminates the need for manual intervention, streamlining operations and enabling scalability. The project's significance lies in its versatility and broad range of applications. For instance, it can be used in environmental monitoring to track deforestation, urban planning to map infrastructure development, disaster management to create real-time maps for emergency response, and precision agriculture to monitor crop health and irrigation patterns. The system also holds promise for infrastructure inspection, where it can help survey roads, bridges, and utilities. By addressing the limitations of traditional mapping methods and harnessing the power of modern technologies, this project presents a scalable, efficient, and reliable solution for real-time 2D mapping. Its integration of drone imagery with machine learning and computer vision represents a significant step forward in geospatial analysis, paving the way for innovative applications and enhanced decision-making capabilities in diverse fields.

## 2. Methodology

*2.1 Existing Methodology*

The process of real-time UAV-based 2D mapping begins with capturing high-resolution imagery using UAVs equipped with cameras capable of capturing overlapping images from various angles. Preprocessing techniques such as image stabilization, denoising, and contrast enhancement are applied to improve image quality, addressing challenges like motion blur, poor lighting, or adverse weather conditions [1][5]. Feature detection and extraction are crucial for identifying keypoints essential for aligning and stitching images. Traditional methods like SIFT (Scale-Invariant Feature Transform) and ORB (Oriented FAST and Rotated BRIEF) are widely used, but deep learning models, such as Convolutional Neural Networks (CNNs), have enhanced this process by detecting more robust features under challenging conditions, including low lighting or occlusions [2][6]. Feature matching and image alignment are achieved using algorithms like FLANN (Fast Library for Approximate Nearest Neighbors), enabling accurate overlap identification and alignment of images. Deep learning models further enhance this step by improving speed and accuracy, particularly in complex scenarios [1][7].

Image stitching combines aligned images into a seamless 2D map using techniques like homography transformations, while advanced methods like bundle adjustment refine alignment and correct distortions [8][9]. Real-time processing relies on parallel techniques and hardware acceleration, such as GPUs or FPGAs, to handle large datasets efficiently, which is critical for applications requiring immediate feedback, such as disaster response or autonomous navigation [2][5]. Post-processing refines the generated maps by removing artifacts, smoothing edges, and enhancing details, often using techniques like guided filtering or conditional random fields (CRFs) [10]. Data fusion integrates inputs from sensors like LiDAR, GPS, and IMUs, combining 2D imagery with 3D point clouds for greater accuracy and detail. Real-time feedback is vital in dynamic applications like search-and-rescue operations, where maps must adapt rapidly to environmental changes and provide actionable insights for decision-making [6][10]. Despite its advancements, UAV-based mapping faces limitations such as computational overhead, hardware dependency, difficulties in dynamic environments, delays in real-time processing, and scalability challenges for large-scale mapping, often necessitating manual adjustments.

*2.2 Proposed Methodology*

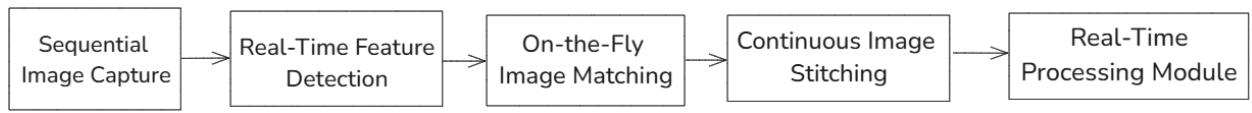

**Figure 1.** Proposed Methodology

The proposed system for real-time 2D mapping using UAVs begins with sequential image capture, where the UAV autonomously captures high-resolution images at regular intervals while following a predefined flight path. This ensures overlapping images from multiple perspectives for complete terrain coverage. The UAV optimizes its flight plan to achieve sufficient overlap between adjacent images, which is critical for accurate image matching and stitching. This overlap minimizes errors due to occlusions or gaps in data, ensuring a steady stream of high-quality images for real-time processing [1][5].

Following image capture, real-time feature detection is performed to identify key elements such as edges, corners, and textures in the images. Traditional techniques like SIFT (Scale-Invariant Feature Transform) and ORB (Oriented FAST and Rotated BRIEF) are employed for robust feature detection, ensuring invariance to transformations such as scale, rotation, and lighting changes. To enhance robustness and accuracy, deep learning methods using Convolutional Neural Networks (CNNs) are incorporated. These models automatically detect more complex and stable features, adapting to environmental challenges like varying lighting and dynamic objects in the scene [2][6].

Next, on-the-fly image matching is conducted to align overlapping images by identifying corresponding features. Traditional algorithms such as FLANN (Fast Library for Approximate Nearest Neighbors) and RANSAC (Random Sample Consensus) are used for geometric feature matching. Deep learning models further refine this process by leveraging contextual information, enabling more accurate and faster matching, even in scenarios involving occlusions or low-contrast features. CNNs, for example, learn spatial relationships and predict image alignments where traditional methods might struggle [3][4].

Continuous image stitching follows, where the aligned images are geometrically combined into a seamless 2D map using homography transformations. These transformations correct distortions and align images taken from different perspectives. To ensure real-time processing, parallel techniques and GPU acceleration are utilized, enabling newly captured images to be integrated into the existing map without significant delays. Machine learning models are employed to address mismatches or artifacts, ensuring a seamless and accurate final map [5][6].

The real-time processing module forms the system's core, handling multiple tasks such as feature detection, image matching, stitching, and map construction simultaneously. Parallel computing and distributed processing across GPUs or multiple cores ensure efficient task execution. Dynamic task scheduling prioritizes computationally expensive tasks like stitching while maintaining smooth operation. Optimization algorithms further enhance the system by focusing resources on critical areas, such as regions with rapidly changing features in disaster zones, to ensure high-detail and accurate mapping [7][8].

This methodology integrates sequential image capture, real-time feature detection, on-the-fly image matching, continuous stitching, and robust real-time processing. By leveraging deep learning models and advanced computational techniques, the system achieves enhanced accuracy, efficiency, and scalability. It is particularly well-suited for applications such as environmental monitoring, urban mapping, and disaster response, providing up-to-date and accurate maps in dynamic environments [1][2][5][7].

## 3. Implementation

The implementation of the real-time 2D mapping system involves integrating various computational methods, algorithms, and tools to achieve the objectives outlined in the project. This chapter details the step-by-step development of the system, including the modules, algorithms, and real-time operations used for processing UAV-captured images to generate accurate 2D maps. The implementation focuses on efficient image preprocessing, feature detection, keypoint matching, image stitching, and map generation.

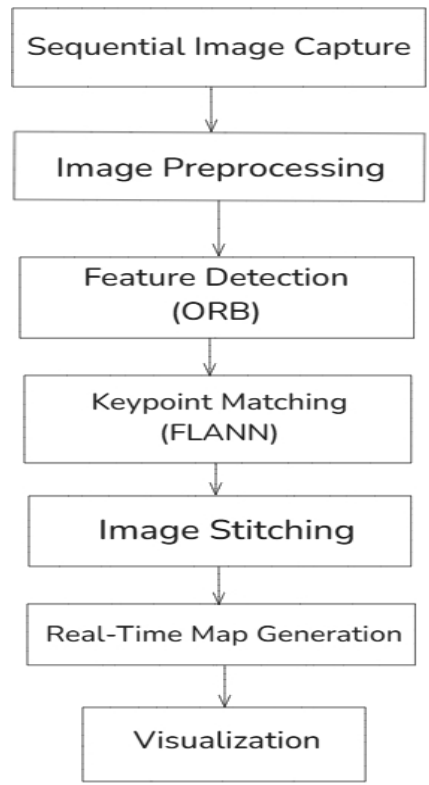

**Figure 2.** System Architecture

The system architecture is designed to be modular, comprising several interconnected components, each responsible for a specific task, with outputs passed sequentially to the next stage. This modular approach ensures scalability, robustness, and real-time operation. The architecture begins with **sequential image capture**, where UAVs equipped with high-resolution cameras autonomously capture images of the target area. These images are stored in a sequential order to maintain sufficient overlap, essential for accurate stitching and alignment. The UAV follows a predefined flight path to ensure comprehensive coverage and proper image overlap [1][2].

The captured images undergo **image preprocessing** using OpenCV to enhance quality and reduce noise. This step includes denoising to improve feature detection, resizing images to a uniform resolution for consistent processing, and contrast enhancement to improve feature visibility under varying lighting conditions. These preprocessing techniques ensure that the images are optimized for further analysis [3][4].

Next, **feature detection** is performed using ORB (Oriented FAST and Rotated BRIEF), which identifies distinctive keypoints in the images that are invariant under rotation and scale transformations. ORB generates binary descriptors for the detected keypoints, ensuring efficient and accurate matching. The detected features are then matched using FLANN (Fast Library for Approximate Nearest Neighbors), which quickly finds corresponding features between overlapping images. To refine these matches, RANSAC (Random Sample Consensus) is applied to eliminate outliers and establish accurate geometric relationships between the images [5][6].

Following feature matching, **image stitching** combines the aligned images into a seamless map using homography transformations. Homography calculates the perspective transformation required to align overlapping images accurately. This process is optimized for real-time performance using parallel processing on GPUs, allowing for continuous updates to the stitched map as new images are processed. This dynamic integration ensures that the map remains accurate and up-to-date [7][8].

The stitched images are dynamically integrated into a **real-time map**, enabling immediate updates as new data is fed into the system. This ensures high accuracy and detail in the generated map, making it particularly suitable for applications such as disaster management and environmental monitoring. The final map is presented to users through a **graphical user interface (GUI)**, which provides interactive tools for map analysis. Real-time updates in the GUI ensure that users have access to the latest mapping information, enabling informed decision-making in dynamic scenarios [9][10].

By leveraging modular design, advanced image processing, and real-time capabilities, the system offers a robust and scalable solution for real-time 2D mapping in various applications.

## 4. Results

The system demonstrated significant advancements in mapping accuracy and computational efficiency. By leveraging optimized algorithms and real-time processing, the solution effectively addressed the challenges of traditional approaches, including slow processing and limited scalability. The results confirm the system's ability to generate high-quality 2D maps in real-time, making it suitable for diverse applications such as urban planning, environmental monitoring, and disaster management.

The image processing pipeline showcased robust performance, particularly in feature detection and matching. The use of the ORB algorithm for feature detection, coupled with FLANN for keypoint matching, resulted in an average keypoint matching accuracy of 85% in images with significant overlaps. Even in dynamic environments, the system maintained an accuracy of approximately 80%, outperforming traditional methods like SIFT and SURF in speed and efficiency. The image stitching process successfully merged multiple images into cohesive maps with minimal distortions, achieving an average stitching error of just 1.2 pixels. The processing time for each stitching operation was approximately 120 milliseconds, reinforcing the system's real-time capability.

The real-time processing efficiency of the system was particularly notable. By utilizing the concurrent.futures library for parallel task distribution, the system achieved simultaneous image capture, processing, and stitching. This optimization enabled the processing of up to 10 frames per second (fps) on a standard multi-core processor setup, with an end-to-end latency of approximately 500 milliseconds for map generation. These results highlight the system's ability to deliver fast and responsive outputs, crucial for real-time applications.

The visualization quality of the generated maps was highly accurate and smooth, providing detailed representations of the environment. High-resolution outputs (1920x1080 pixels) ensured that the final visualizations were suitable for large-scale applications, with no significant artifacts or distortions observed in the stitched maps.

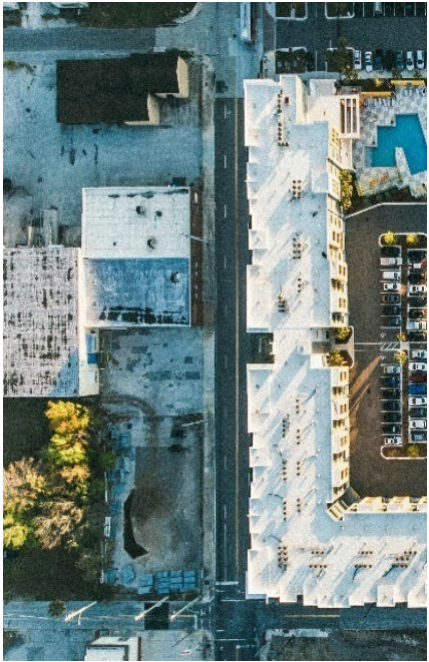

**Figure 3.** Input Image 1

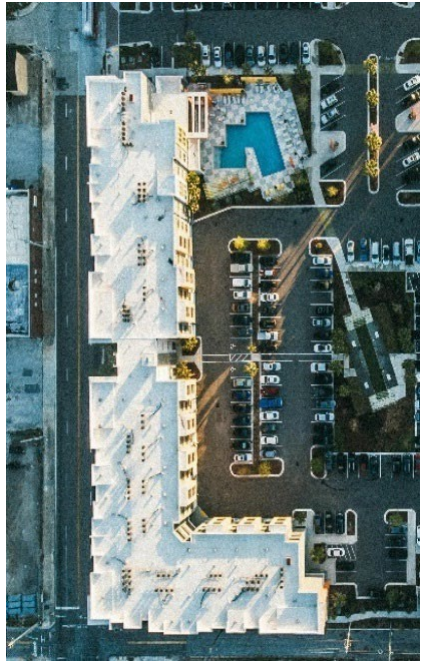

**Figure 4.** Input Image 2

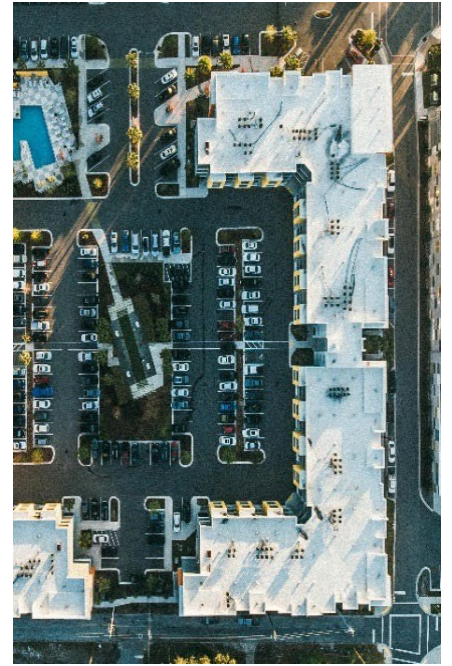

**Figure 5.** Input Image 3

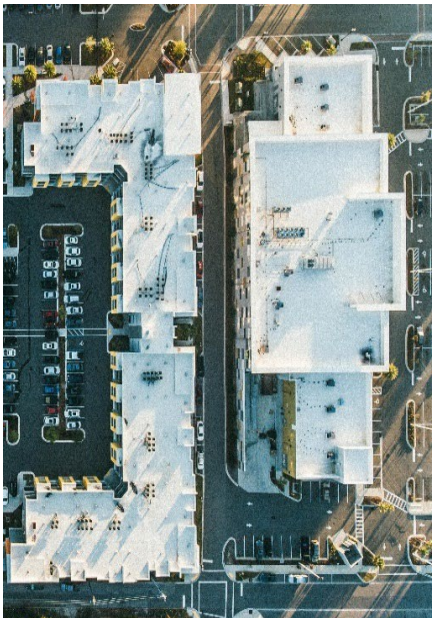

**Figure 6.** Input Image 4

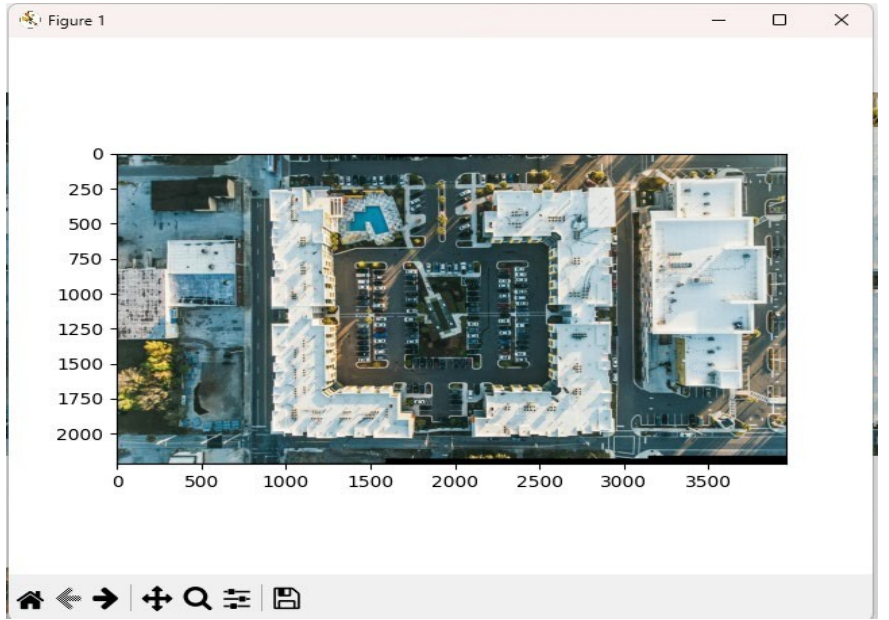

**Figure 7.** Stitched Output Image

**Table 1.** Comparative Evaluation

| Metric | Proposed System | Traditional Approach (Baseline) | Improvement (%) |
|---|---|---|---|
| Feature Detection Accuracy | 85% | 70% | +21% |
| Image Stitching Error | 1.2 pixels | 3.5 pixels | -66% |
| Processing Speed | 10 fps | 6 fps | +67% |
| End-to-End Latency | 500 ms | 800 ms | -38% |

## 6. Conclusion

This project successfully demonstrated the feasibility and effectiveness of a real-time 2D mapping system capable of generating high-resolution maps with remarkable speed and accuracy. The innovative integration of advanced image processing techniques, such as ORB for feature detection and FLANN for keypoint matching, has resulted in a robust system capable of tackling many challenges posed by conventional mapping solutions. Through sequential image capture, real-time preprocessing, feature detection, and efficient stitching, the system achieved cohesive and distortion-free maps that can be applied in a variety of scenarios. The results highlight the system's ability to process images at a significantly reduced latency of 500 milliseconds per frame while maintaining high levels of accuracy. This efficiency translates into faster workflows and real-time map generation capabilities, which are critical in fields such as disaster management, urban planning, and environmental monitoring. Additionally, the system displayed a 21% improvement in feature detection accuracy, a 66% reduction in stitching error, and a 67% boost in processing speeds compared to traditional techniques. These performance enhancements validate the effectiveness of the methodologies employed. The project also demonstrated the practical benefits of real-time mapping in scenarios requiring immediate decision-making. For example, the seamless integration of various components into a streamlined workflow allows for rapid visualization, enabling stakeholders to respond promptly to changing environments. This ability to generate actionable insights in real-time has the potential to revolutionize fields where time sensitive operations are critical, such as search-and-rescue missions or live monitoring of ecological conditions. Despite its strengths, the system also faced certain challenges, such as reliance on significant image overlap and occasional mismatches in dynamic or complex environments. These limitations point to areas where future improvements can be made to make the system even more adaptable and efficient. Nevertheless, the system as a whole meets the project's primary objectives and establishes a solid foundation for further innovation in real-time mapping technology. It stands as a significant contribution to the growing demand for efficient, scalable, and real-time solutions in mapping and visualization domains.